# Morphology Decoder: Untangling Heterogeneous Texture and Determining Permeability with Machine Learning 3D Vision

**Omar Alfarisi, Aikifa Raza, Djamel Ozzane, Mohamed Sassi, Hongtao Zhang, Hongxia Li, Khalil Ibrahim, Hamed Alhashmi, Hamdan Alhammadi, and TieJun Zhang**

## Abstract:

Permeability has a dominant influence on the flow behavior of a natural fluid, and without proper quantification, biological fluids (Hydrocarbons) and water resources become waste. During the first decades of the 21st century, permeability quantification from nano-micro porous media images emerged, aided by 3D pore network flow simulation, primarily using the Lattice Boltzmann simulator. Earth scientists realized that the simulation process holds millions of flow dynamics calculations with accumulated errors and high computing power consumption. Therefore, accuracy and efficiency challenges obstruct planetary exploration. To efficiently, consistently predict permeability with high quality, we propose the Morphology Decoder. It is a parallel and serial flow reconstruction of machine learning-driven semantically segmented heterogeneous rock texture images of 3D X-Ray Micro Computerized Tomography (μCT) and Nuclear Magnetic Resonance (MRI). For 3D vision, we introduce controllable-measurable-volume as new supervised semantic segmentation, in which a unique set of voxel intensity corresponds to grain and pore throat sizes. The morphology decoder demarks and aggregates the morphologies' boundaries in a novel way to quantify permeability. The morphology decoder method consists of five novel processes, which we describe in this paper, these novel processes are (1) Geometrical: 3D Permeability Governing Equation, (2) Machine Learning: Guided 3D Properties Recognition of Rock Morphology, (3) Analytical: 3D Image Properties Integration Model for Permeability, (4) Experimental: MRI Permeability Imager, and (5) Morphology Decoder (the process that integrates the other four novel processes).

## Introduction

Permeability has a dominant influence on the flow properties of natural fluids. And computer vision is of extensive use for permeability determination (*1*), for heterogeneous Cretaceous carbonate (~110 million years ago) (*2, 3*), aided by the 3D pore network concept (*4-8*). Geoscientists simulate fluid flow through porous media morphology (*9, 10*), primarily using Lattice Boltzmann (*11*), making the results depend heavily on the accuracy of pore size (*12*) and pore connectivity (*13-15*) images. Three hindering challenges faced pore network Lattice Boltzmann simulators (*16, 17*), simulation assumption limitations (*18, 19*), high computation power when targeting better accuracy or using a more massive data set (*19, 20*), and the resolution of the images used for extracting the pore network (*21-24*). Researchers have also used deep learning, specifically the convoluted neural network (CNN) algorithm (*25-27*). Although CNN identified pore size distribution, we observed limited application to 1D nuclear magnetic resonance data or 2D μCT images. A 2D image analysis of heterogeneous Cretaceous morphology produces a localized description (*28*) and does not represent the whole 3D morphological structure. None of the vendors delivered an acceptable permeability value when testing four commercial computer vision models (*29*) using the 2D image for permeability determination. The use of CNN for pore size identification and permeability prediction from 3D μCT is progressing (*30*); however, its focus is on homogenous sandstone (*30, 31*). On the contrary, Cretaceous carbonate is heterogeneous due to fossils (we also use the term “bioclast” interchangeably in this paper) contents. At the same time, diagenesis impacts Cretaceous carbonate texture (*32*). Carbonate heterogeneity made it complicated for CNN to analyze its morphology compared to sandstone (*31*).

In our research, to overcome the challenges of pore network simulation and solve heterogeneous porous media properties quantification challenges, we introduce Morphology Decoder (MorphD). It is an interdisciplinary computer vision-guided physics for fluid flow properties quantification. MorphD is a standalone technique – non-simulation based – predicts permeability using 3D vision (*21*). MorphD ensembles image processing, machine learning, 3D printing, 3D μCT, and MRI vision. Rather than the conventional interpore connectivity (*33-35*) of the pore network, MorphD builds a pore throat network (PorThN) by segmenting 3D μCT intensity of comparable morphology using machine learning algorithms (*36, 37*). Then MorphD identifies – with MRI intensity (*38, 39*) – the impact of pore throat size (*PorTS*) on the control volume boundary of the 3D segmented section (*40-42*).

In MorphD, we propose an inventive scheme, the Controllable-Measurable-Volume (CMV), to substitute the Representative Elemental Volume (REV) concept (*31, 43-48*). MorphD finally rebuilds all 3D segmented sections like a LEGO (*49, 50*) to predict – without any fluid flow simulation (*51-54*) – the 3D permeability (*55-57*), using two parallel and serial aggregation governing equations (*58*). MorphD validates the equation's results with polymer-based 3D printed micromodel experiments (*21, 59, 60*) (Fig. 1).

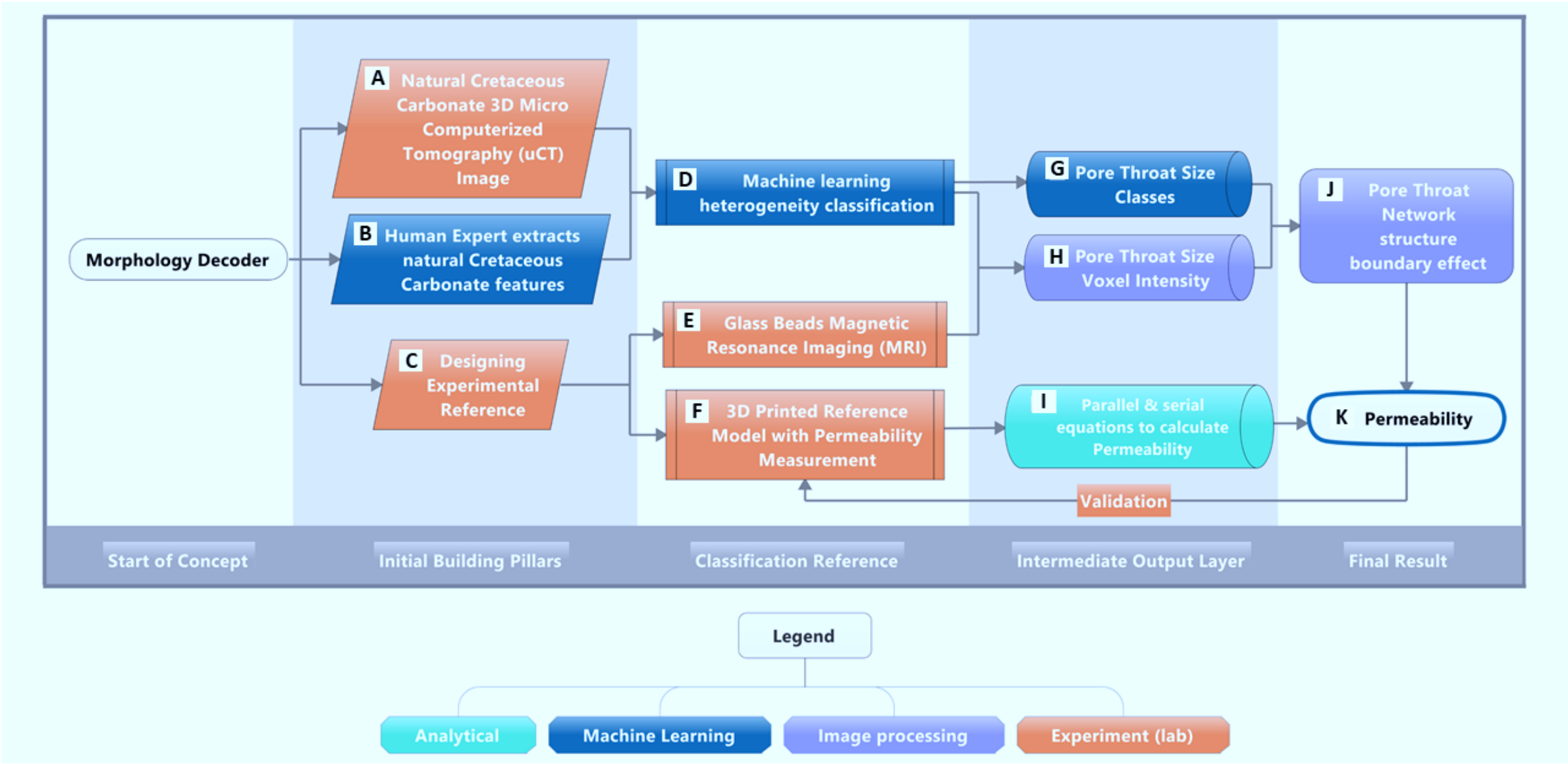

***Fig. 1. Morphology Decoder. (A)*** *High-resolution 3D micro computerized tomography (µCT) image is the target of our machine learning guided image analysis for determining permeability.* ***(B)*** *Human expert extracts feature from the µCT image and label them for the machine to learn from them.* ***(C)*** *To calibrate the Magnetic Resonance Imaging (MRI) model for pore throat size determination, we designed two types of 3D Micro Models (3DMM).* ***(D)*** *As per human expert training labels, the machine classifies Cretaceous carbonate to the desired groups.* ***(E)*** *Glass Beads 3D Micro Model (GB3DM) references pore throat sizes corresponding to MRI image voxel intensity.* ***(F)*** *Printed Mesh 3D Micro Model (PM3DM) produced the series and parallel governing equation verification to determine heterogeneous morphology permeability.* ***(G)*** *Machine learning prediction result of pore throat sizes classification is one of the three inputs to calculate permeability.* ***(H)*** *Voxel intensity is the 2$^{nd}$ input to the calculation of permeability.* ***(I)*** *The final governing equations for calculating 3D permeability.* ***(J)*** *The structure boundary effect of the pore throat network, which we also call a controllable, measurable volume (CMV), represents a discrete permeability value of each classified morphology inside the Cretaceous carbonate.* ***(K)*** *By using the CMV and the governing equations, we calculate Permeability.*

## Morphology Decoder Method

### 1- Geometrical – Novel 3D Permeability

In homogenous texture, permeability depends on *PorTS*, while in heterogeneous one (*38*), permeability depends on the pore throat network. We define a pore throat network (PorThN) as a heterogeneous texture with multi pore throat sizes distributed in a specific system to produce a unique morphology. In exploring research options for identifying pore throat sizes and PorThN, we identified five possible paths: analytical, machine learning, image processing, experimental, and simulation (Fig. 2). We focused on the geometrical analysis (*61-64*). We target understanding the physics of the main factors that derive permeability. Machine learning provides an efficient and consistent quality process in segmenting a mass of 3D image data (*65, 66*). Image processing eases the interpretation of machine learning outcomes to produce usable quantitative results (*67, 68*). While experiments provide a calibration and validation assurance of our analytical, machine learning, and image processing approaches. We omitted the use of simulation to free researchers from simulation dependency; instead, we quantify permeability with MorphD.

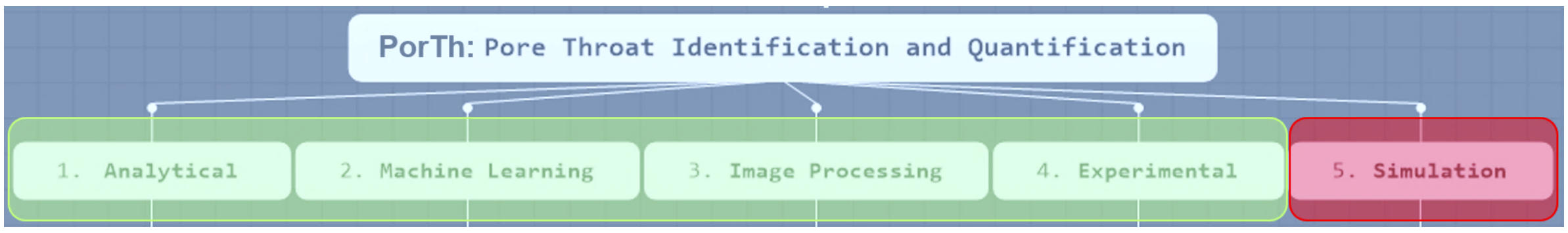

***Fig. 2. Pore throat identification options.*** *Green highlighted boxes are research approaches we considered, while the red highlighted box is a research option that we omitted in our research.*

The analytical path focuses on the 2D and 3D geometrical analysis of objects. It describes the grain and the texture to determine *PorTS* and network. We start first with a fundamental step of analyzing a homogenous geometry. Then we add further complexities to analyze heterogeneous morphology, which contains more than one pore geometry. Finally, we apply the learning to analyze natural Cretaceous carbonate. Intergranular and intragranular pores (*69*) are two different geometrical systems that Cretaceous carbonate comprises.

We start from the homogenous cubic structure of well-sorted spherical grains displayed in 2D projection (*38, 70-73*) (Fig. 3). In Fig. 3, we see that the edges of the circle are also the edges of the pore throat. We deduced Axiom-0; a pore throat is a plane with enclosure from all directions.

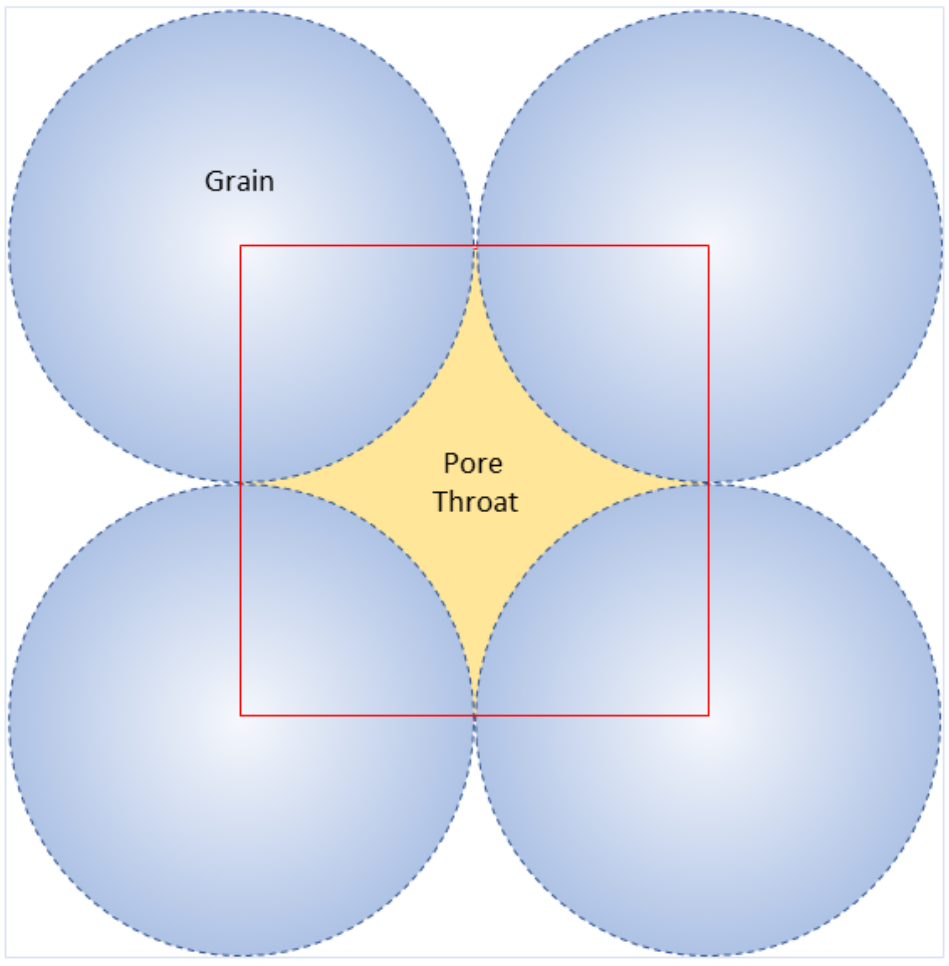

***Fig. 3. Handcrafted 2D projection of four spherical grains representing one of the most homogeneous sphere-based configurations.*** *The red square, which connects the centers of the circles, represents the largest perpendicular plane to the center of the four rings. This plane act as the starting point for image processing. The red square is in the X-Y plane, and the image analysis moves perpendicular to this plane. The red square also represents the largest plane of the pore. The pore size is the square lateral, equivalent to a circle diameter. The yellow concaved diamond is the pore throat. Imagine the red square and yellow concaved diamond shape in 3D; it is an arched pyramidal structure. The arch pyramid has a square base, four sides that are one-eighth of a sphere, and a concaved diamond top.*

We show the fundamental geometrical analysis in Eq. 1, determining the pore area in 2D at the largest plane. The corners of this plane are the four points of contact of the eight spheres:

$$Pore\ Size\ Area = Small\ Red\ Square\ Area\ of\ (Fig. 3) = 4r_g^2 \tag{1}$$

where,

$r_g$ : The grain size (grain radius).

In Eq. 2, we show the area determination of *PorTS* in 2D (plane view of Fig. 3):

$$PorTS = Small\ Red\ Square\ Area\ -\ Circle\ Area = 4r_g^2 - \pi r_g^2 = 0.858r_g^2 \tag{2}$$

The relation between the area of pore size and *PorTS* in 2D is the ratio between Eq. 1 and 2:

$$\frac{Pore\ Throat\ Size}{Pore\ Size} = \frac{0.858r_g^2}{4r_g^2} = 0.2146 \tag{3}$$

The natural grain-based porous system is of 3D configuration, not 2D, and this requires calculating the volume rather than the area. In Fig. 4, A, we see the same porous system of Fig. 3, but in 3D. The full 3D representation of the pore and pore throat is in Fig. 4, B.

Porosity is the void volume ratio to an object's total bulk volume (void + solid), so it is dimensionless property (volume/volume). Porosity does not change with the change of grain size. Still, it varies only with the grain configuration (i.e., sorting and compaction), determining the ratio transformation between the void and the total bulk volume (*38*). The porosity of our Cretaceous sample ranges from 0.18 to 0.32. This range gives an average porosity of 0.25 for the pore structure of rhombohedral configuration, as shown in Fig. 4, C (*38, 61*). It is also called rhombohedral-pyramidal (*74*), and in this paper, we call it rhombohedral. The porosity value does not change with grain size change if the grain configuration remains the same, Fig. 4, D.

The *PorTS* changes with grain size change, Fig. 4, D. In carbonate, porosity has no proportional relation with *PorTS*. Therefore, porosity has no direct relationship with the rock types for rock type classification, but this is not the case in clastics (Sand-Shale environment), where porosity can directly relate to rock types. Also, in carbonate, the porosity has limited influence on fluid flow (i.e., permeability), while *PorTS* and *PorThN* have the dominant impact. While in clastics, the total porosity is not relatable to permeability, while the effective porosity has a significant effect on permeability. The total and effective porosity in carbonate is the same value due to the extremely limited or non-existence of shale in carbonate rock, except in the carbonate rock containing non-connected vugs.

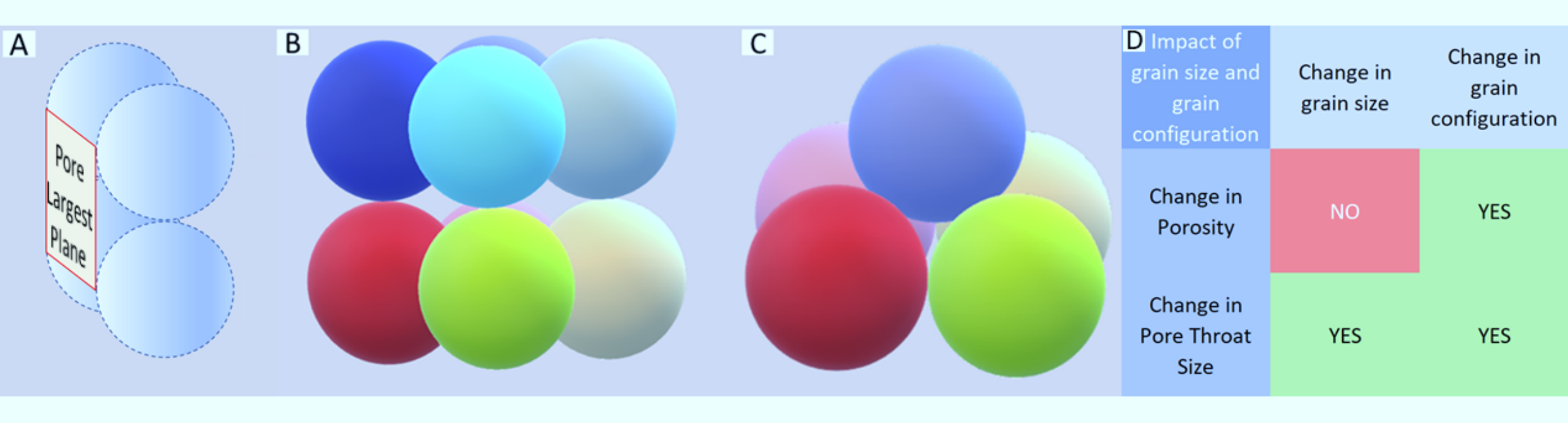

***Fig. 4. Different 3D configurations of spherical grains. (A)*** *Four spherical grains in 3D represent homogeneous well-sorted configuration. The red square represents the pore size. Each square lateral is equivalent to the grain diameter.* ***(B)*** *Eight spherical grains represent a well-sorted cubic configuration. Even though the cubic configuration does not represent natural grain distribution, we use it to analyze the geometrical equations, then move to a more complex structure.* ***(C)*** *The rhombohedral system better represents the original configuration of poorly sorted grains in nature.* ***(D)*** *The impact of grain size and layout on porosity and pore throat size.*

The grain size impacts the pore throat radius, Fig. 4, D, also shown in Eq. 2. We derive the equation for the rhombohedral configuration in Fig. 4, C, to determine *PorTS*. We start with a more straightforward structure than rhombohedral, triclinic, shown in Fig. 5, A, a 2D illustration of three spheres. We derived the equations that govern the pore size and *PorTS* below from Fig. 5, A, and B:

$$Pore\ Size = Red\ Triangle\ Area = \frac{\sqrt{3}}{4} r_g^2 \tag{4}$$

$$PorTS_{Triclinic} = 4 * Green\ Tringle\ Area - 0.5 * Circle\ Area = 4\frac{\sqrt{3}}{4} r_g^2 - \frac{\pi}{2} r_g^2 = 0.162 r_g^2 \tag{5}$$

We notice that Eq. 5 represents a 2D $PorTS_{Triclinic}$, while our desired geometrical configuration is a 3D rhombohedral configuration (Fig. 4, C), which holds more complexity than triclinic (Fig. 5, A). The 3D cubic configuration of eight spheres shown in Fig. 4, B consists of six faces: top, bottom, and four slides. A pore throat shape of a concaved diamond on each face, like the yellow area shown in Fig. 3. Therefore, the 3D pore throat area of cubic configuration is the sum of six concaved diamonds areas, Eq. 2, to be $5.148{r_g}^2$.

Then we calculate the Effective 3D Pore Throat Size of cubic configuration ($PorTS_{cubic_{3D}{}_{Effective}}$) as shown in Eq. 6 below:

$$PorTS_{cubic_{3D}{}_{Effective}} = \frac{A_{cubic_{PorT}}}{N_{PorT} \cdot N_{C\forall_{inets}}} r_g^2 = \frac{5.148}{6 * 2} r_g^2 = 0.429 r_g^2 \tag{6}$$

where,

$A_{cubic_{PorT}}$ : The area of all pore throats of cubic configuration,

$N_{PorT}$ : The number of pore throats in a 3D configuration,

$N_{C\forall_{inets}}$ : The number of outlets of the fluid flow control volume.

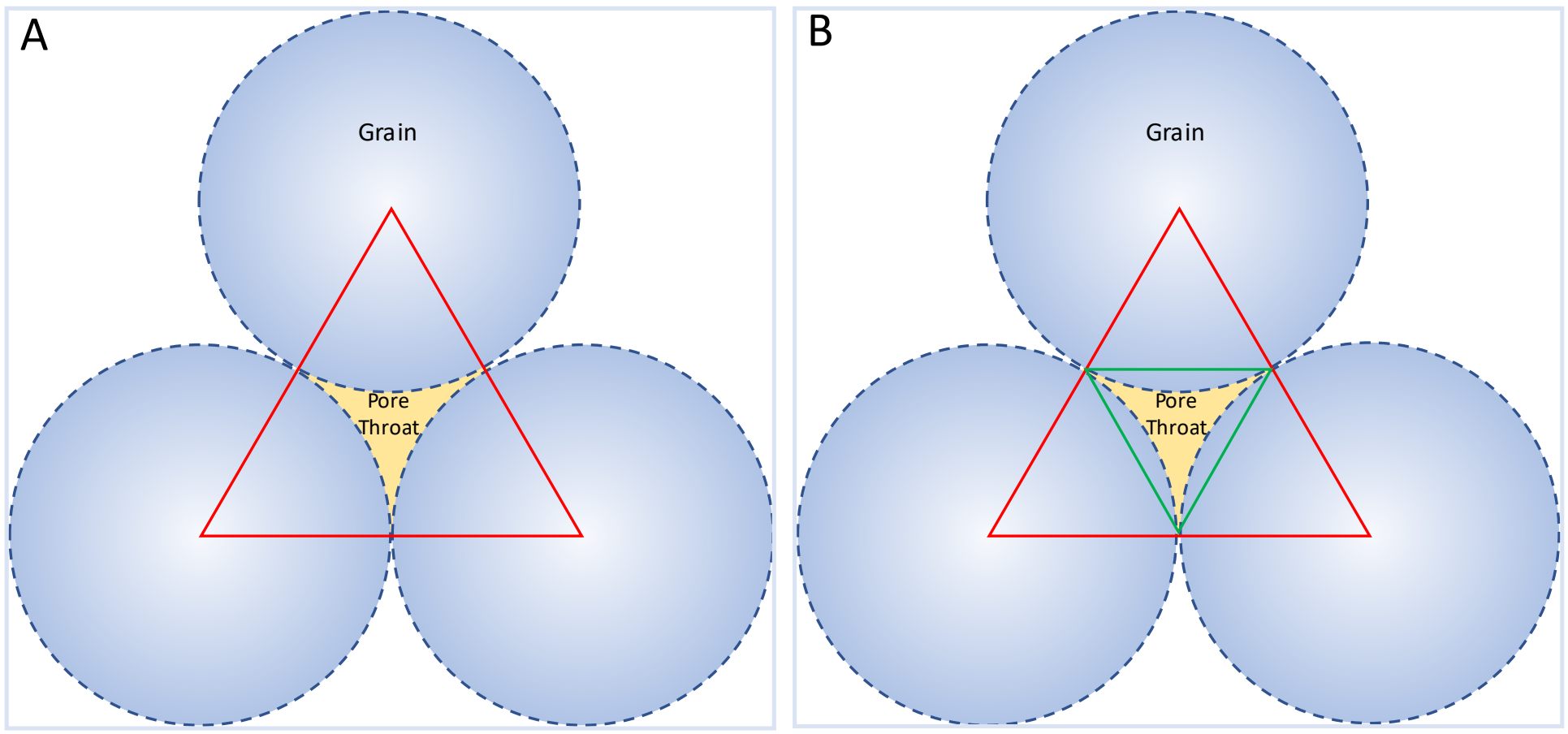

***Fig. 5. Triclinic configuration of spherical grains. (A)*** *The red triangle shows the pore size, while the yellow concaved triangle shows a pore throat area, and both – pore size and pore throat size – depend on grain size, while porosity does not depend on grain size but grain configuration.* ***(B)*** *Geometrically analyzed triclinic configuration to derive the PorTS equation, Eq. 5.*

The 3D triclinic configuration of eight spheres consists of six faces that hold two different shapes of pore throats; the top, bottom, and two sides have a pore throat shape of a concaved diamond. The other two sides hold a concaved triangle pore throat shape: two-pore throats per side. Therefore, the 3D pore throat area of triclinic configuration is the sum of four concaved diamonds and four concaved triangles, Eq. 2 and 5 to be $4.08{r_g}^2$. Then we calculate the Effective 3D Pore Throat Size of triclinic configuration ($PorTS_{triclinic_{3D\,Effective}}$) as shown in Eq. 7, below:

$$PorTS_{triclinic_{3D\,Effective}} = \frac{A_{triclinic_{PorT}}}{N_{PorT}\,.\,N_{C_{\forall inets}}} r_g^2 = \frac{4.08}{8*2} r_g^2 = 0.255 r_g^2 \qquad (7)$$

where,

$A_{triclinic_{PorT}}$ : The area of all pore throats for triclinic configuration.

The 3D rhombohedral configuration of eight spheres consists of six faces that hold two different shapes of pore throats; the top and bottom faces hold a pore throat shape of a concaved diamond. The four sides have a concaved triangle pore throat shape: two-pore throats per side. Therefore, the 3D pore throat area of rhombohedral configuration is the sum of two concaved diamonds and eight concaved triangles areas, Eq. 2 and 5, to be $1.716{r_g}^2$.

Then we calculate the Effective 3D Pore Throat Size of rhombohedral configuration ($PorTS_{rhombohedral_{3D}{}_{Effective}}$) as shown in Eq. 8 below:

$$PorTS_{rhombohedral_{3D}{}_{Effective}} = \frac{A_{rhombohedral_{PorT}}}{N_{PorT} \cdot N_{C_{\forall inets}}} r_g^2 = \frac{1.716}{10 * 2} r_g^2 = 0.0858 r_g^2 \quad (8)$$

where,

$A_{rhombohedral_{PorT}}$ : The pore throats area of rhombohedral configuration.

We rewrite Eq. 8 in terms of grain surface area, as shown in Eq. 9 below:

$$PorTS_{rhombohedral_{3D}{}_{Effective}} = A_{surface_{grain}} = 0.02731 \pi r_g^2 \quad (9)$$

Permeability is a resultant of both grain size and grain configuration to form a proportional relation between permeability and grain surface area (*58*) " this physical aspect of permeability has been used to create empirical equations for prediction of permeability," as described below in Eq. 10:

$$Permeability\ (mD) = A_{surface_{grain}}\ (\mu m^2) \quad (10)$$

For 3D rhombohedral configuration, we determine the permeability by substituting Eq. 9 and Eq. 10 to produce Eq. 11 as shown below:

$$k_{3D_{rhombohedral}} = PorTS_{rhombohedral_{3D}}{}_{Effective} = 0.0858r_g^2 \quad (11)$$

We display Eq. 11, with experimental data (*39*) on different grain sizes and permeability with poorly sorted grains (rhombohedral), as shown in Fig. 6.

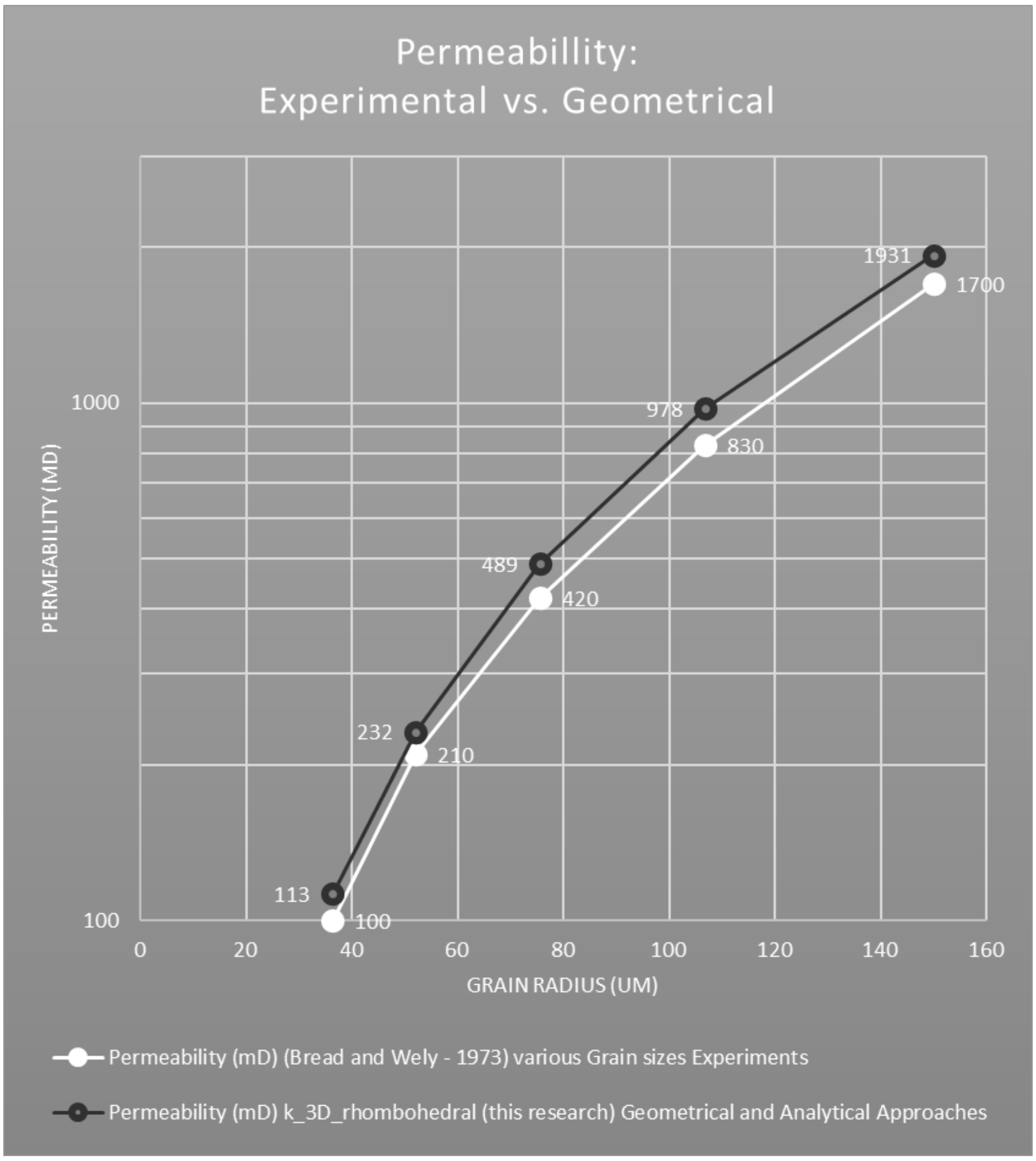

***Fig. 6. Experimental and analytical permeability determination.*** *The validation of permeability determination for rhombohedral configuration using an analytical equation derived from grain radius compared to experimental permeability values.*

## 2- Machine Learning – Novel 3D Properties Recognition of Carbonate Morphology

The main morphology decoder deliverable is the permeability of heterogeneous Cretaceous rock *($k_{HeC}$)*. One of the cornerstones of morphology decoder for determining $k_{HeC}$ is Eq. 11. Another critical cornerstone is differentiating heterogeneity zones (DHZ), where machine learning-based computer vision plays the role of deciding DHZ for classifying minerals (*75*) with the 3D μCT (*21, 76*) images. In Cretaceous carbonate, more than one mineral exists, and the volumes of these minerals are quantifiable (*77*), in our case, calcite and pyrite (*21*). Fig. 7, A, shows the original μCT image with 28 um resolutions. To identify DHZ, we run machine learning with the Random Forest (RF) algorithm to perform image recognition of different rock sections (*21*). After conducting comparisons between several algorithms, we chose the RF algorithm because RF proved to be the most suitable one with the highest accuracy (*78*). Fig. 7, B, shows the training image with four critical features – DHZs. We display the classification results in Fig. 7, C; with Pyrite (red color), Open Vugs (Green color), Intergranular-1 (Pink color), and Intergranular-2 Bioclast (Yellow color). We separate each DHZ geometrically as a discrete block to create the controllable-measurable-volume (CMV) as shown in Fig. 7, D – G. These CMV's are Pyrite, Open Vugs, Intergranular-1, and Intergranular-2 Bioclast, respectively. Each CMV has a specific grain size with a permeability value – $k_{3D_{rhombohedral}}$ – that aggregates to produce the Cretaceous rock permeability.

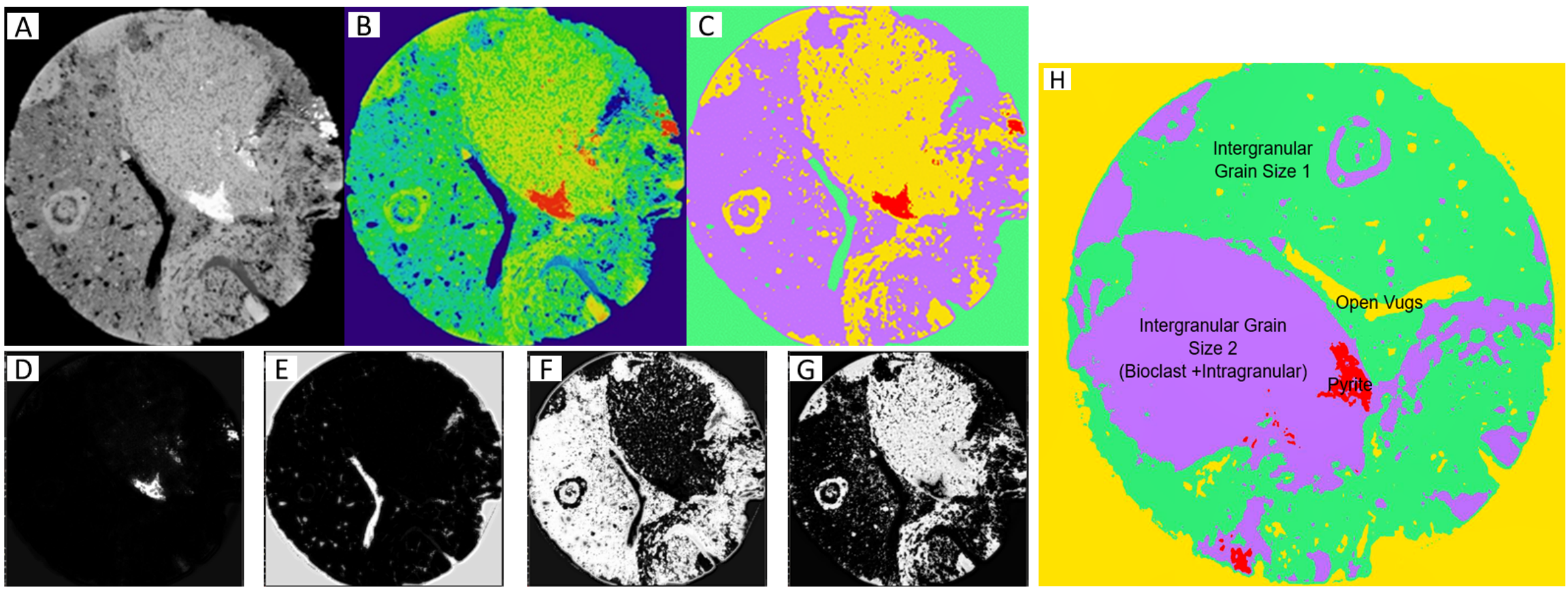

***Fig. 7. Machine learning guided 3D properties recognition of natural rock morphologies. (A)*** *The original 3D µCT image with 28 um resolution shows we can identify different gray shades representing different morphology. (B) The training outcome of the machine learning guided image processing to recognize the different zones in Cretaceous carbonate. Each different morphology has a different impact on the flow property of a natural fluid. We can see four differentiating heterogeneity zones (DHZ) recognized by the machine. Each of these DHZs volume boundaries becomes an attribute to define the controllable-measurable-volume (CMV) we use to reconstruct the rock for permeability determination.* ***(C)*** *The results of the machine learning segmentation of DHZs* ***(D)*** *DHZ of Pyrite are separated, as shown in white color shade.* ***(E)*** *DHZ of Open Vugs is split, as shown in white color shade.* ***(F)*** *DHZ of Intergranular-1 is separated, as shown in the white color shade.* ***(G)*** *DHZ of Intergranular-2 is split, as shown in white color shade.* ***(H)*** *Morphologies have labels corresponding to their DHZ.*

## 3- <u>Analytical – Novel 3D Vision Property Integration Model for Permeability</u>

We innovated a new permeability aggregation process for the 3D image stack. We aggregate the permeability using parallel and serial permeability equations (*58*) shown in Eq. 12 and 13 (Fig. 8, A) to produce the general 3D permeability equation of heterogeneous rock, $k_{HeC}$. We call this aggregation; 3D Property Integration Model (3DPIM). The steps for achieving 3DPIM starts keeping the flow direction (the arrows in Fig. 8, B) perpendicular to the Control Volume ($C_\forall$) boundary. In the next step of 3DPIM, we integrate the Permeability of each 2D slice (x-y plane), using the parallel permeability equation, Eq. 13, to calculate the slice permeability. The last step of 3DPIM is integrating the z-axis direction for the 3D stack using the serial equation, Eq. 12, Fig. 8, C.

$$k_{avg_{serial}} = \frac{l_1+l_2}{\left(\frac{l_1}{k_1}\right)+\left(\frac{l_2}{k_2}\right)} \tag{12}$$

$$k_{avg_{parallel}} = \frac{h_1k_1+h_2k_2}{h_1+h_2} \tag{13}$$

where,

$l_1, l_2$ : the length of the section,

$h_1, h_2$ : the height of the section,

$k_1, k_2$ : the permeability of each CMV.

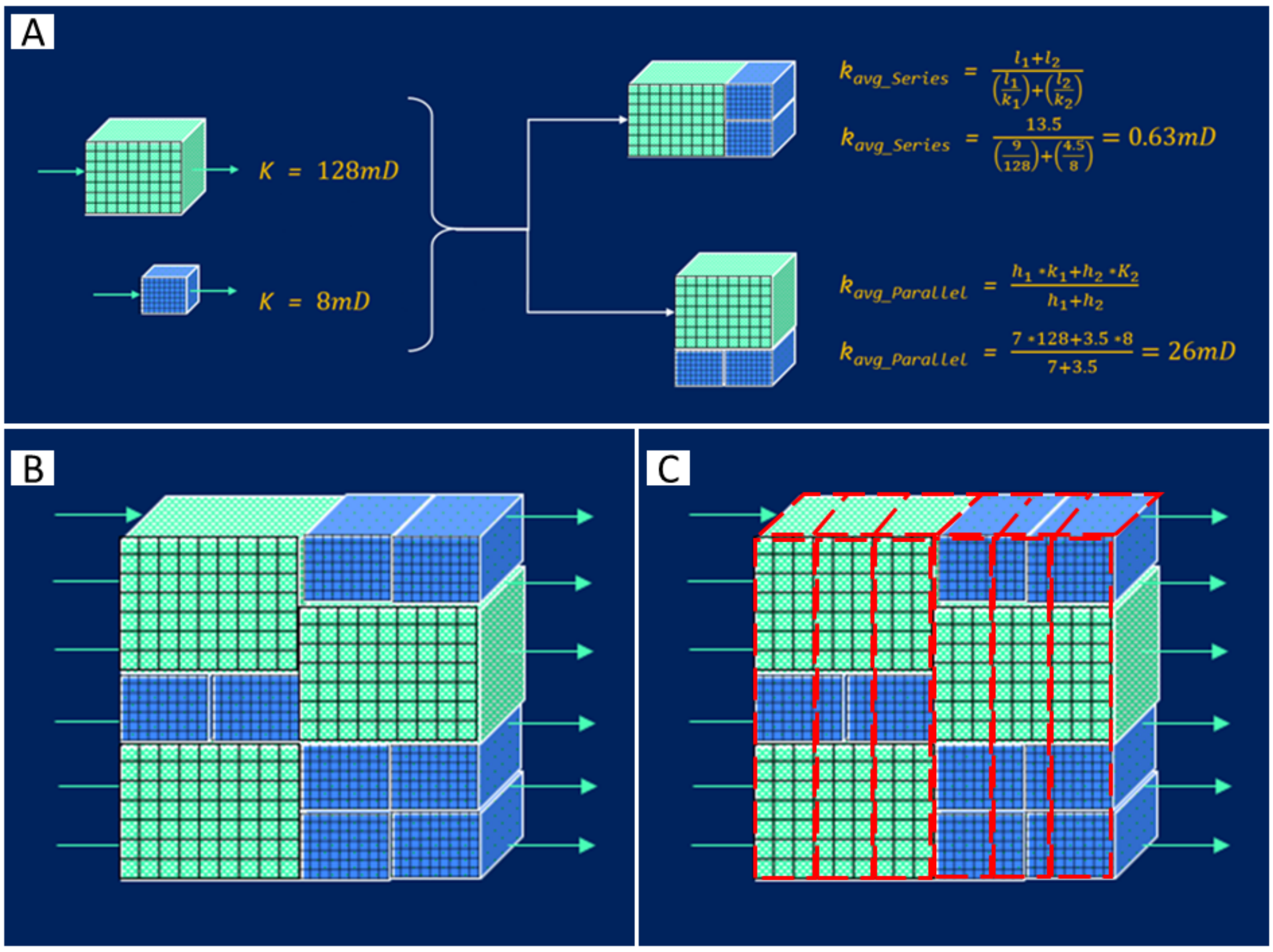

***Fig. 8. The Morphology Decoder Process. (A)*** *Two examples of applying parallel and serial equations.* ***(B)*** *3D Permeability Model visualization.* ***(C)*** *3D Permeability determination based on control volumes corresponds to 2D slices of 3D image stacked in the z-axis.*

To validate our novel 3DPIM model, we designed a 3D porous media (3DPM) conceptual model, shown in Fig. 9 top (Conceptual Design). We draw five different 3DPM configurations, utilizing two mesh sizes (mesh inner laterals): 2000 um (the dark blue color mesh shown in Fig. 9) and 4000 um (the green color mesh shown in Fig. 9). These configurations reflect homogeneity and heterogeneity with two types of homogenous rocks – Sample 1 and 2 – shown in Fig. 9, one serial configuration of two homogenous rocks – Sample 3 – shown in Fig. 9, one parallel configuration of two homogenous rocks – Sample 4 – shown in Fig. 9, and heterogeneous rock of arbitrary distribution of two rock types – Sample 5 – shown in Fig. 9. The inner laterals sizes demonstrate a double difference in size; this helps us accurately differentiate the flow effects. The outer dimensions of 3DPM cylinders are 7.8 cm x 3.8 cm (length x Diameter), a size that can fit the flooding apparatus we use for permeability measurement. The conceptual model had its final 3D drawing engineered with Computer-Aided Design (CAD) software, as shown in Fig. 9, the second from the top (3D Computer Design). Then we 3D-printed 3DPM five cylinders with a polymer material to have the final physical look of the cylinders shown in Fig. 9, the second from the bottom (3D Printed). Then we measured the five 3DPM for determining permeability by equipping each cylinder with a rubber sleeve and injecting air at 35 psi (~241 KPa) sleeve conformance pressure. The conformance pressure we used is the maximum pressure the 3DPM can hold before deformation occurs to the cylinders. We also calculated the permeability using Eq. 12 and 13. Then we validated the calculated permeability with the measured permeability, as shown in Fig. 9, the bottom (Samples Measured and Predicted), to prove our novel 3DPIM model's ability to quantify permeability with high confidence.

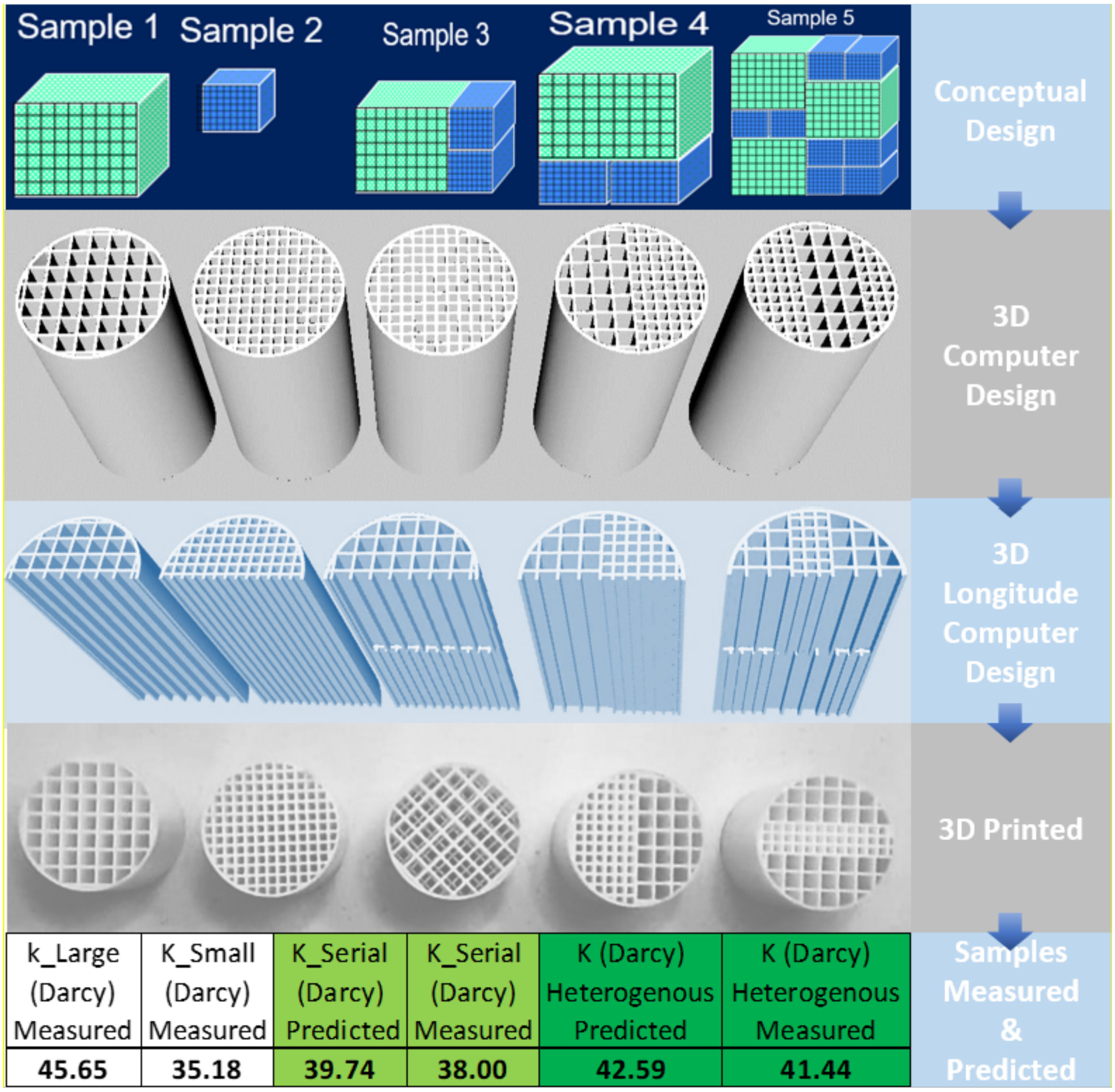

***Fig. 9. The 3D aggregation model validation with 3D printed samples of various mesh sizes and configurations.*** *We compare the predicted (calculated using our novel 3DPIM model) and the physically measured permeability.*

## 4- <u>Experimental – Novel MRI Permeability Imager</u>

We proposed MRI as a grain radius quantifier. MRI can sense the fluid (containing hydrogen) rather than anything else. Nuclear magnetic relaxation time (T2) value differs for different pore sizes. In MRI measurement, the larger the pore, the slower the T2 (*79*). Despite all the progress in NMR and MRI technology, there has not been any published work that shows a direct relation between MRI Image Intensity (MRIII) and grain size with rhombohedral configuration until writing this paper. Earlier in this paper, we proved the direct relation between pore size and throat size. We also demonstrated the direct link between pore throat size and grain size of rhombohedral configuration. Then we established the direct relation between grain size and permeability of rhombohedral configuration. Therefore, we targeted the ability to measure the grain size using MRIII for the first time and human history in this research part. We built a new laboratory apparatus specifically for this purpose. We used an NMR system of 0.5 Tesla. The MRI instrument setup consists of three main parts, the magnetic field with the core holder, the NMR radio frequency and temperature controller, and the 3-phase pressurized flooding system, as shown in Fig. 10.

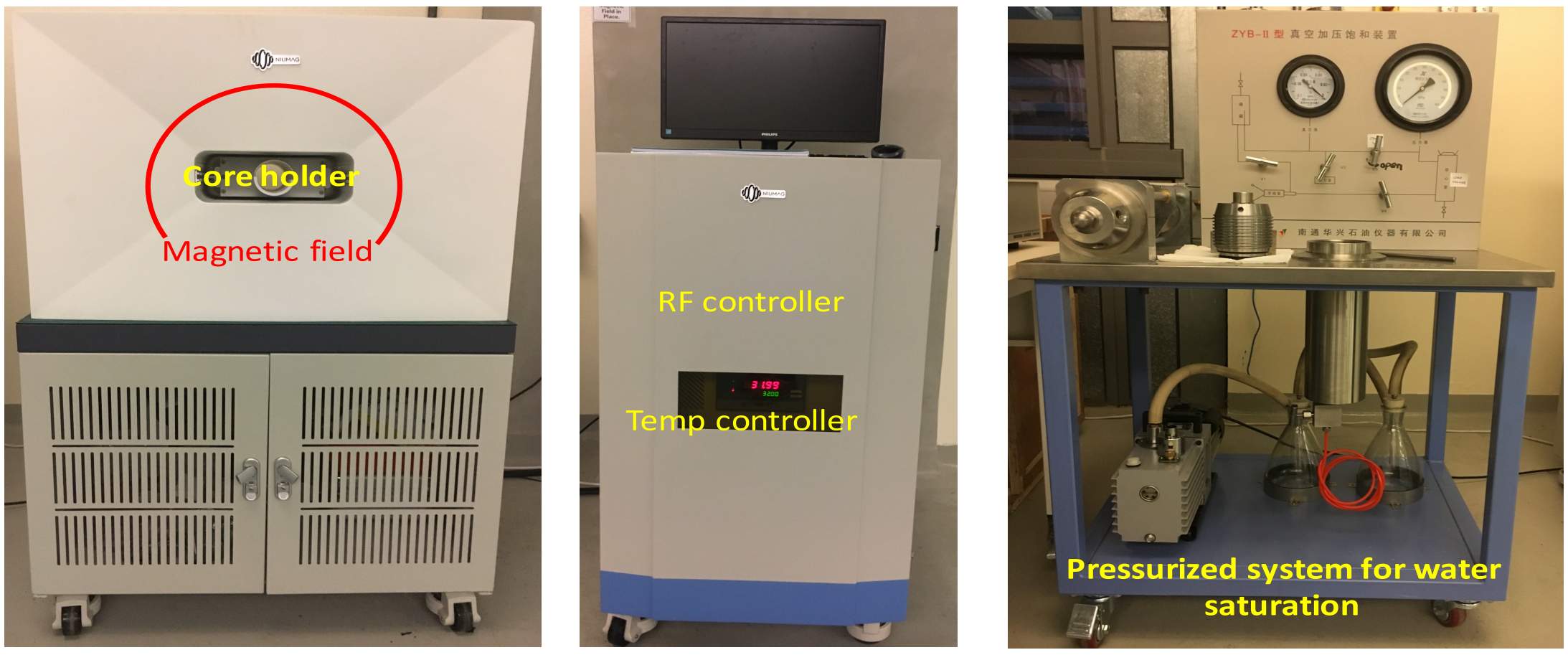

***Fig 10. Bench Top NMR and MRI Analyzer with Pressurized Flooding System.***

To prove our hypothesis of quantifying the effect of grain size poorly sorted configuration (rhombohedral) on MRI signal, we used glass beads of various grain sizes – five sizes – packed poorly. The dimension of the glass beads we used are of five ranges: 520 um – 700 um, 200 um – 300 um, 70 um – 125 um, 8um – 50 um, and 1 um – 7 um. We acquired MRI images for the five different sizes, as shown in Fig. 11. The measurement results have confirmed our hypothesis of MRIII's ability to establish a direct link to the grain size of the rhombohedral configuration.

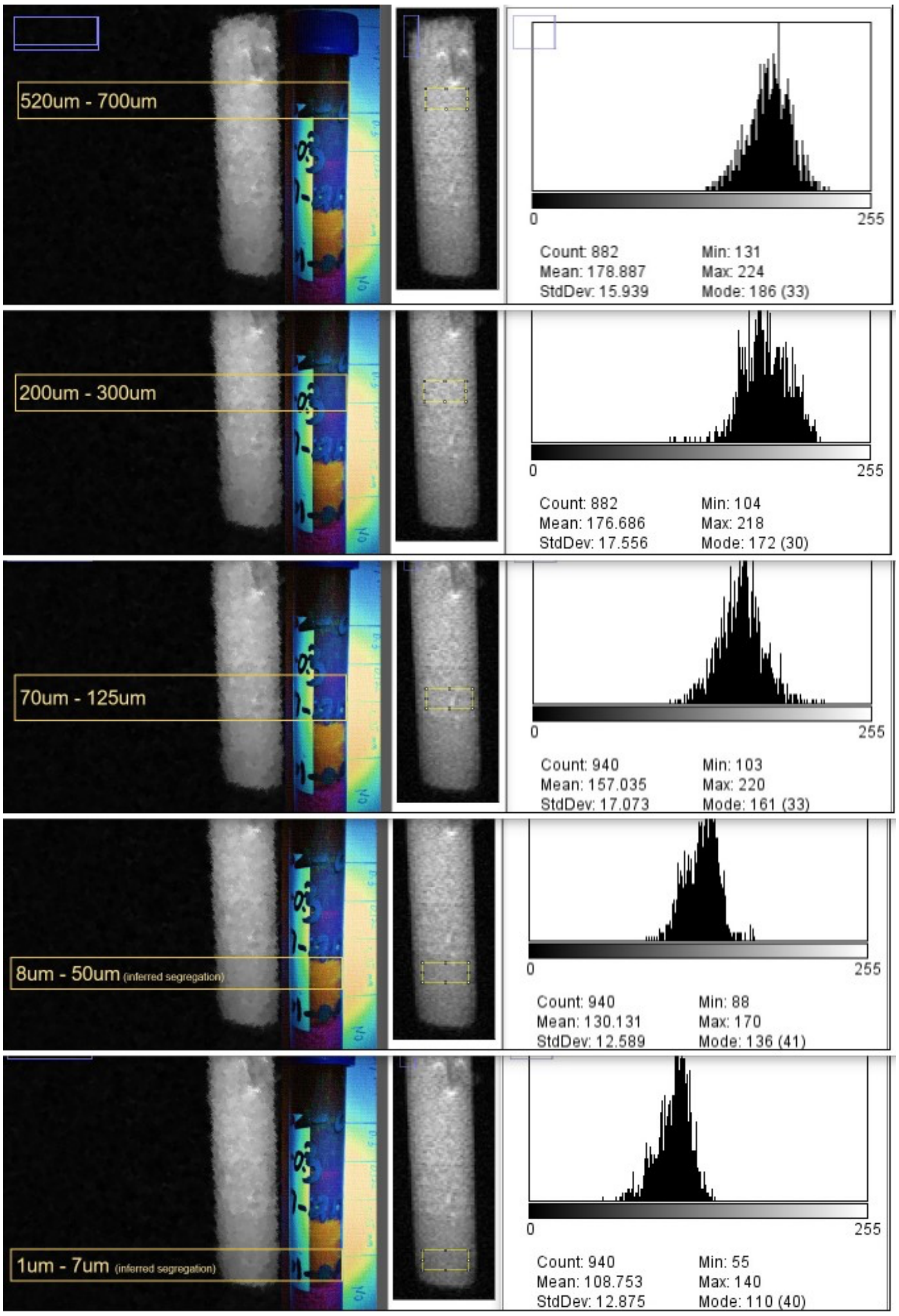

***Fig. 11. Glass beads with various sizes measured with the bulk (x-y-z) MRIII.***

We then calibrate MRIII quantification of grain size with rhombohedral configuration. This step is vital to building a novel model that enables humanity to measure grain size using MRIII. We select a narrower range of MRIII measurement (x-y axis) for three distant sizes of glass beads: 1500 um, 400 um, and 50 um, (Fig. 12).

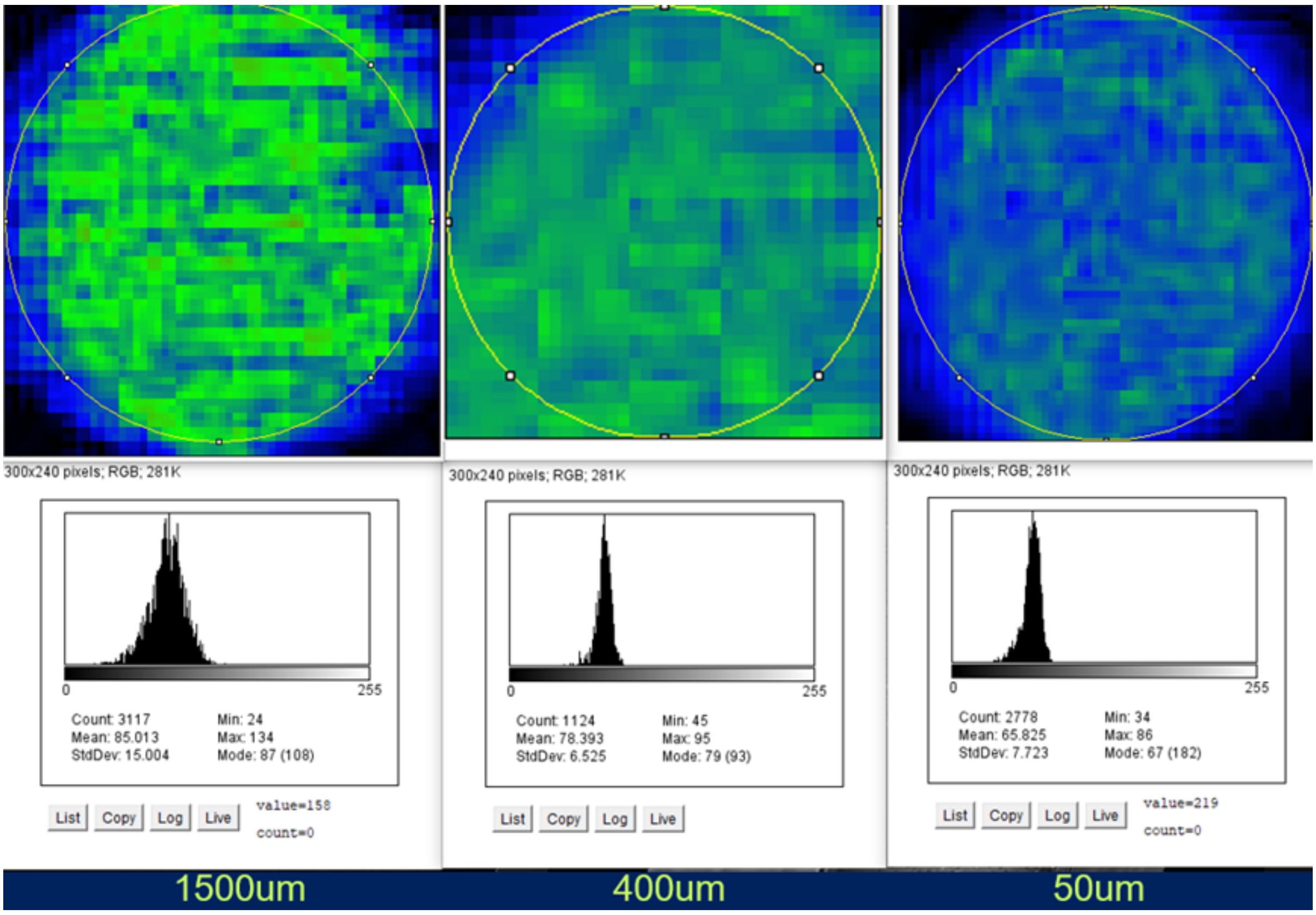

***Fig. 12. MRI Image Intensity (MRIII) calibration for quantifying grain size using three sizes of Glass Beads.***

Then based on the fine selection of glass beads and MRIII, we build the MRI-Grain Calibrator Model (MGCM), as shown in Fig. 13. To construct the MGCM model and enable future use of MGCM consistently, we maintained all the acquisition parameters for the MRI machine to be unchanged at all the stages of the experimental work.

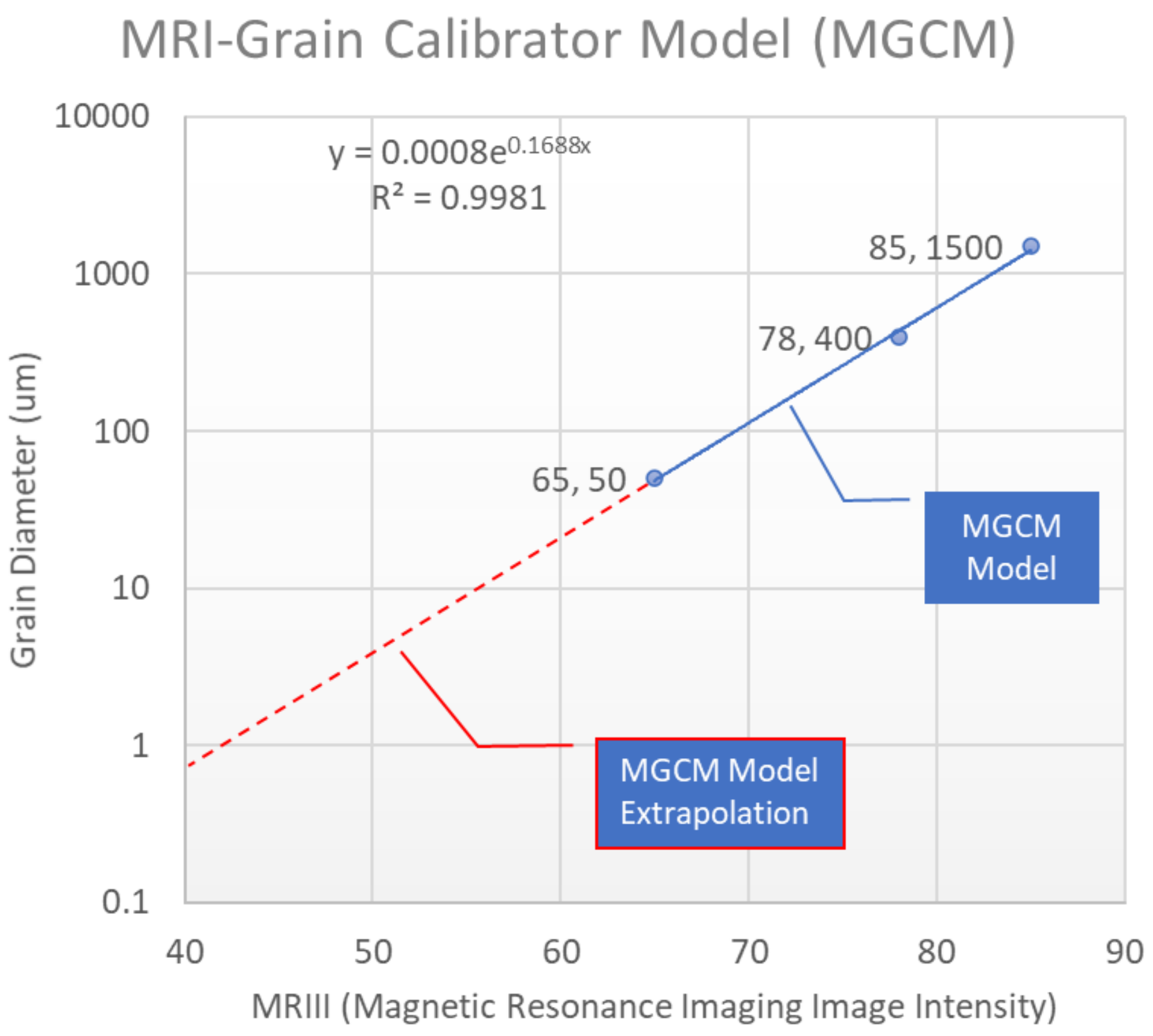

***Fig. 13. MRI-Grain Calibrator Model (MGCM).*** *The model shows the direct relation between MRIII Image Intensity (MRIII) and Grain Diameter. The MGCM model is in blue, while we displayed an extrapolation of the MGCM for illustration to show the potential of further calibration.*

To apply and validate the MGCM model on a natural rock, we acquired MRIII on Cretaceous carbonate rock, as shown in Fig. 14. As mentioned earlier, the acquisition parameters of the MRI device are the same as the ones used for acquiring the MGCM calibration model. Consistency between calibration model acquisition parameters and subsequent measurements implies that if the MRI device has a different physical configuration, different running energy, or the distance between MRI sensors and measured objects is different, and another calibration must be obtained. Therefore, for the usage of Morphology Decoder in field well drilling and planetary exploration, each calibration setup provides an MGCM chart update. The MGCM calibration models the fluid type that saturates the rock. The fluid type surrounds our novel

Permeability Imaging Logging Tool (PILT), interchangeably called Permeability Imager, including an NMR, MRI, and µCT measurement device. While for well depth measurement, a Gamma-Ray (GR) measuring device – an available tool in the market - is an additional integrated part of the PILT. The calibration includes modeling the effects of the thickness of the drilling fluid mud-cake formation on the drilled wells' walls that affect the µCT imaging focusing. Therefore, the wellbore diameter measurement with a caliper tool and the drilling fluid composition and salinity are part of fluid and mud-cake correction for MGCM. The calibration of MGCM considers the micro-resistivity log to correct for the fluid type that saturates the rock. These two measurements (caliper and micro-resistivity) – available tools in the market - can be integrated as additional components to PILT.

Although MGCM model values are not universal, the MGCM method processes are all universal. For the natural rock histogram, as for the MGCM model, we record the mean value of the histogram; for this specific rock, MRIII is 67.633. We plug this MRIII value into the MGCM model shown in Fig. 13 to produce a grain diameter value of 68.82 um.

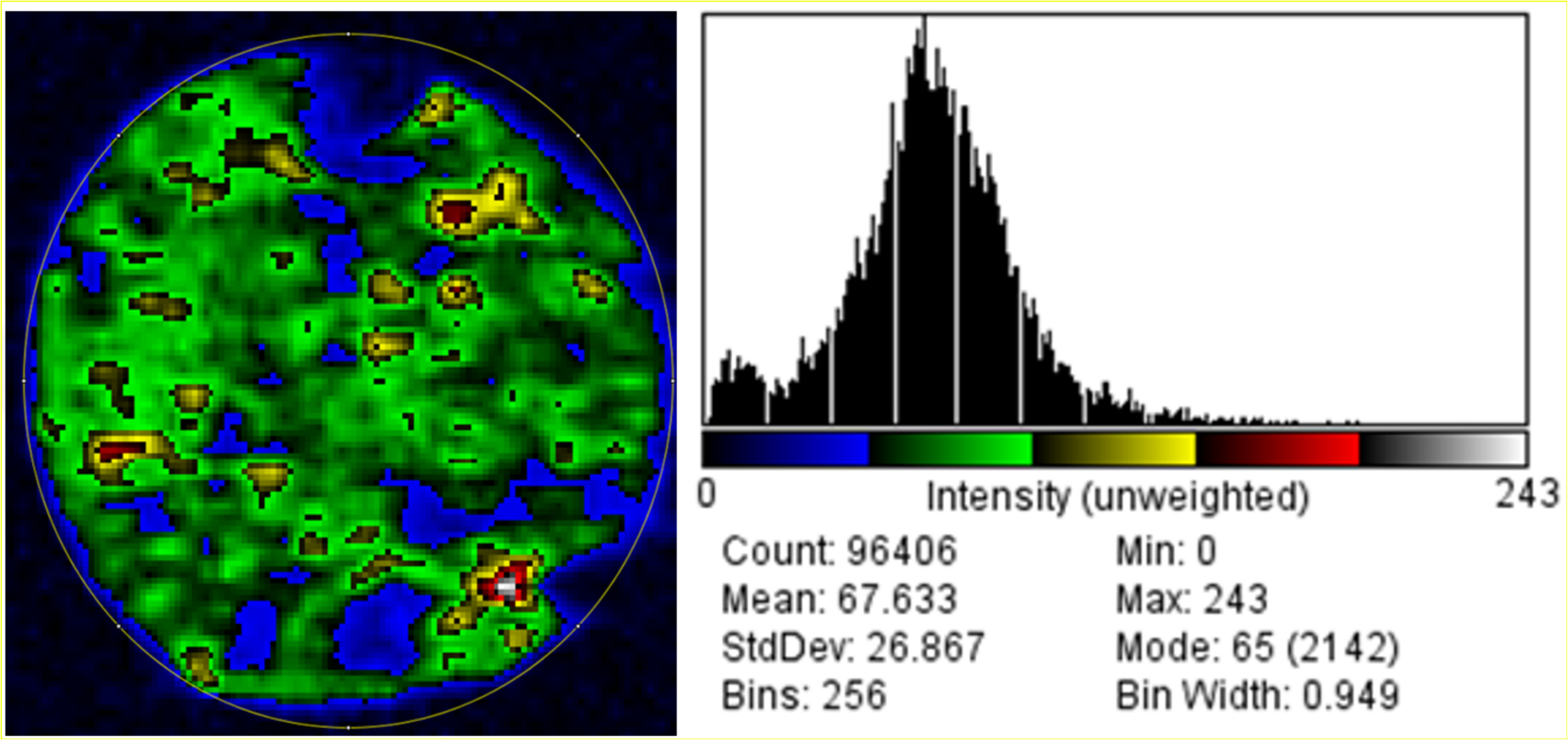

*Fig. 14. MRIII Histogram of natural rock (Cretaceous Carbonate) of Abu Dhabi.*

Then we plug the grain diameter value of 68.82 um on Eq. 11 to calculate $k_{3D_{rhombohedral}}$ for this natural rock sample to find permeability value to be 113 mD. To validate the 113 mD permeability value, which is the result of our innovative process of Morphology Decoder, we have measured the natural rock sample at an independent industry laboratory. The laboratory measurement confirms 126 mD, which is considered a close match to the MorphD permeability value. The petrophysics accepted practice is to have a matching accuracy of less than 0.5 of a logarithmic decade scale. We achieved less than 0.013 of a logarithmic decade scale in our case. We map the Morphology Decoder algorithm (Fig. 15) as a summary road map from imaging to machine learning to CMV to the physics analytical model to experimental calibration and validation.

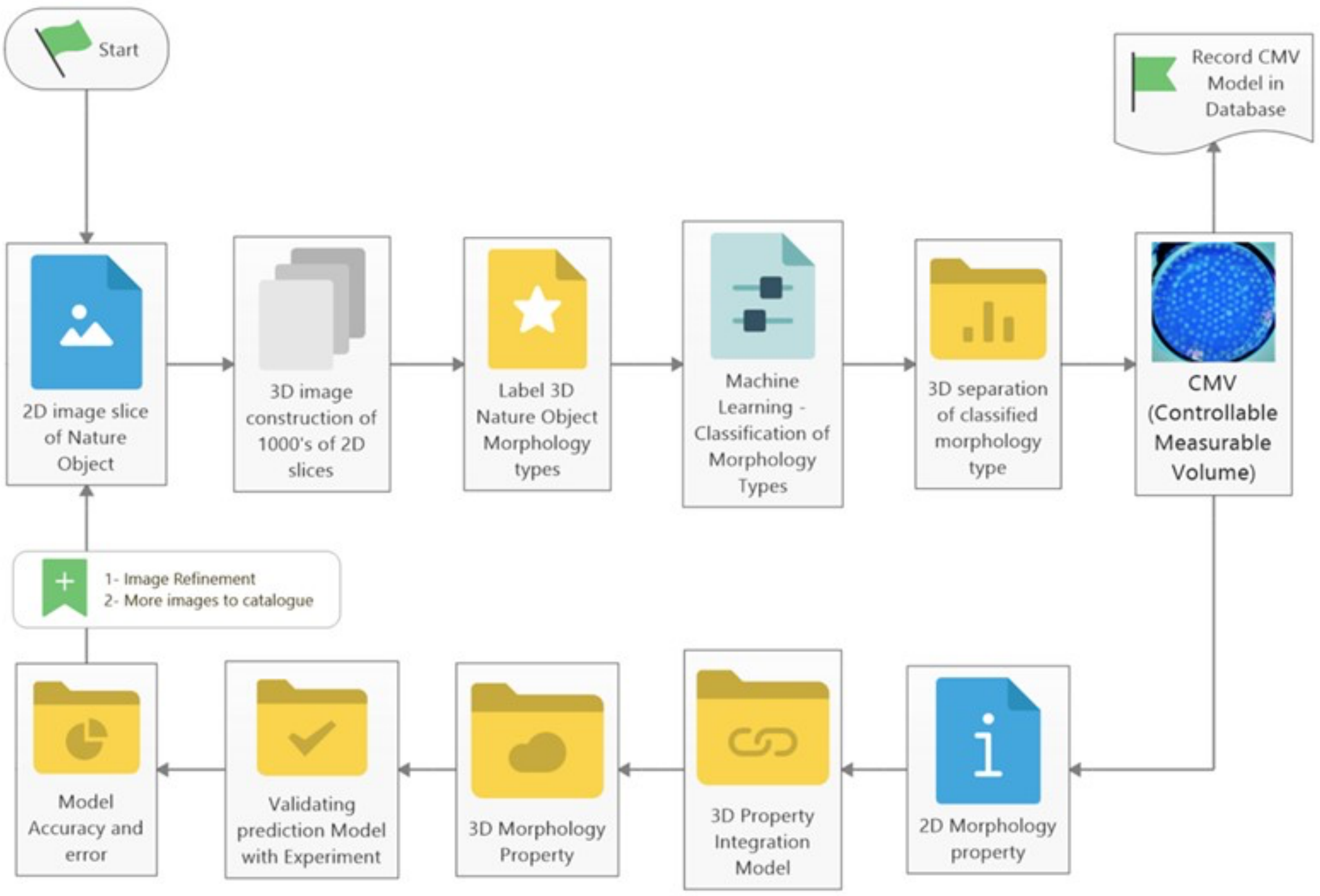

***Fig. 15. Morphology Decoder.*** *A road map shows the novel protocol for quantifying natural rock permeability through building novel Geometrical, Machine Learning, Analytical and Experimental tasks with innovative integration, which we call Morphology Decoder.*

## Conclusion

This research has targeted delivering a novel quantification of simulation-free permeability. An innovative interdisciplinary method, the Morphology Decoder, is the outcome of this research. For developing the Morphology Decoder, we established several interlinks between physical and chemical properties of heterogeneous carbonate rock, supported by geometrical analysis, machine learning, 3D printed calibrator experiments, and analytical derivations, some of which we conclude below:

1- This paper proved several new interlinks between various physical properties:
   a. We have established the link between grain configuration and pore throat size, Eq. 9.
   b. We have established the link between pore throat size and permeability, Eq. 11.

2- Use these established links of (1) above to interpret the images acquired by μCT and MRI to deliver a novel quantification method of natural rock permeability.

3- We created a novel computer vision semantic segmentation representation, Controllable Measurable Volume CMV, guided with calibration models, to have a novel 3D image flow property aggregation, leading to quantifying permeability.

4- Morphology Decoder is an efficient method for developing and interpreting new generation measuring devices. Morphology Decoder-based logging tools (wireline, logging while drilling, logging while coring, permanent sensors, and macro-micro-nano sensing robots (*80*)) is a solution for autonomous planetary exploration in Earth, Moon, and Mars.

## Affiliation

Omar Alfarisi[1,2*], Aikifa Raza[1], Djamel Ozzane[3,4], Mohamed Sassi[1], Hongtao Zhang[1], Hongxia Li[1], Khalil Ibrahim[2], Hamed Alhashmi[2], Hamdan Alhammadi[2], and TieJun Zhang[1].

[1]*Department of Mechanical Engineering, Khalifa University of Science and Technology, PO Box 127788, Abu Dhabi, UAE.*

[2]*Department of USSU Field Development, ADNOC Offshore, PO Box 303, Abu Dhabi, UAE.*

[3]*Directorate of Upstream, ADNOC, PO Box 898, Abu Dhabi, UAE.*

[4]*BP Exploration Operating Co Ltd, BP p.l.c., International Headquarters, 1 St James's Square, London, SW1Y 4PD, UK.*

## Remark

This is a Preprint Paper – which is an updated version attached as Appendix for PCT Patent Application filed at the US Patent Office on Sep 10th, 2021, following a Provisional Patent Application filed at the US Patent Office on Sep 10th, 2020.

## Acknowledgment

The authors appreciate all the support received from ADNOC, ADNOC Offshore, and Khalifa University of Science and Technology in providing the data, enabling laboratory utilization, and other related administrative support. We would like to thank Mr. Yasser Al-Mazrouei, Mr. Ahmed Al-Suwaidi, Mr. Ahmed Al-Hendi, Mr. Mohamed Alghaferi, Mr. Saoud Almehairbi, Mr. Ahmed Al-Riyami, Mr. Andreas Scheed, Mr. Salem Al-Zaabi, and Mr. Mohamed Abdelsalam for their unwavering support, encouragement, and motivation.